\documentclass[runningheads,a4paper]{llncs}

\usepackage{amssymb}
\usepackage{graphicx}

\usepackage{url}
\urldef{\mailsa}\path|soheil.ghafurian@merck.com|

\usepackage{multirow}
\usepackage{hhline}
\usepackage[table]{xcolor}
\usepackage{subfig}

\graphicspath{{Figures/}}

\newif\ifanonymous
\anonymousfalse

\begin{document}

\mainmatter  

\title{Classification of Protein Crystallization X-Ray Images Using Major Convolutional Neural Network Architectures}

\titlerunning{Protein Crystallization Image Classification: Major CNN Architectures}

%
%
\author{
\ifanonymous
anonymous
\else
Soheil Ghafurian$^{1}$
\and 
Peter Orth$^{2}$
\and 
Corey Strickland$^{2}$
\and 
Hua Su$^{2}$
\and
Sangita Patel$^{2}$
\and\\ 
Steven Soisson$^{2}$
\and 
Belma Dogdas$^{1}$
\fi
}
\authorrunning{Protein Crystallization Image Classification: Major CNN Architectures}

\institute{
\ifanonymous
anonymous
\else
$^{1}$Applied Mathematics and Modeling, Global Data Science Center, Merck \& Co. Inc., Rahway, NJ 07065, USA\\
\mailsa\\
$^{2}$Department of Biochemical Engineering and Structure, MRL, Merck \& Co. Inc., Kenilworth, NJ 07033, USA\\
\url{www.merck.com}
\fi
}
%
%

\toctitle{Lecture Notes in Computer Science}
\tocauthor{Authors' Instructions}
\maketitle

\begin{abstract}
The generation of protein crystals is necessary for the study of protein molecular function and structure.
This is done empirically by processing large numbers of crystallization trials and inspecting them regularly in search of those with forming crystals.
To avoid missing the hard-gained crystals, this visual inspection of the trial X-ray images is done manually as opposed to the existing less accurate machine learning methods.
To achieve higher accuracy for automation, we applied some of the most successful convolutional neural networks (ResNet, Inception, VGG, and AlexNet) for 10-way classification of the X-ray images.
We showed that substantial classification accuracy is gained by using such networks compared to two simpler ones previously proposed for this purpose.
The best accuracy was obtained from ResNet (81.43\%), which corresponds to a missed crystal rate of 5.9\%. 
This rate could be lowered to less than 0.1\% by using a top-3 classification strategy.
Our dataset consisted of 486,000 internally annotated images, which was augmented to more than a million to address class imbalance.
We also provide a label-wise analysis of the results, identifying the main sources of error and inaccuracy.

\end{abstract}

\section{Introduction}

Protein crystallography is essential for the study of molecular protein structures.
The function and properties of a protein type is decided by its molecular structure, which is inferred from the protein crystal~\cite{rhodes2010crystallography}.
However the set of chemical conditions conducing to protein crystallization, such as concentration, temperature, precipitant type, and pH are hard to ascertain and different for each protein type.
Determination of appropriate crystallization conditions for a given protein often requires testing many conditions before a successful one is obtained~\cite{luft2007efficient}.
To this purpose, high throughput screening systems are employed to process large numbers of crystallization trials automatically.
These trials need to be periodically inspected in search of forming crystals.
This is a cumbersome task, as the crystallographer needs to go through large numbers of trial images on a daily basis while maintaining a high level of precision in order not to miss crystals.
Therefore, a high precision crystal detection algorithm will substantially advance the field of protein crystallography.

A number of traditional machine learning algorithms have been applied to the protein crystallization problem for feature selection and classification.
Such algorithms include  dynamic programming~\cite{bern2004automatic}, decision trees~\cite{liu2008image}, random forests~\cite{cumbaa2010protein}, Bayesian classification~\cite{hung2014protein}, and multi-layer perceptrons~\cite{sigdel2013real}.
To the best of our knowledge, Yann et al.~\cite{CrystalNet} applied a deep convolutional neural network (CNN) to this problem for the first time.
They used an architecture with four convolutional and three fully connected (FC) layers and called their model CrystalNet.
They reported an accuracy of 90.8\% by applying CrystalNet to the dataset introduced in in~\cite{snell2008establishing2} and~\cite{snell2008establishing1}.

We initially adopted CrystalNet for our dataset and achieved a 10-way testing accuracy of 73.7\%.
We then reduced the number of parameters in CrystalNet roughly in half by adding two max-pool layers and compensated the reduced complexity by increasing the non-linearity of the network through added depth.
The resulting network achieved a testing accuracy of 78.6\%, outperforming CrystalNet while requiring less computation.
The details of this work can be found in\ifanonymous \hspace{.1cm} anonymous \else~\cite{GhafurianISBI}\fi.

The performance gain from the added depth suggests that the model could be improved by the incorporation of higher level features.
Thus, in this work, we apply the more sophisticated CNN architectures from the field of computer vision to the protein crystallization problem.
In the following sections we report the performance of each architecture and provide a detailed analysis of the results for each image label.
We also employ a data augmentation step to tackle the label imbalance, intrinsic to the protein crystallization process, in the the image dataset.

\section{Method}

\subsection{Data acquisition and labeling:}

The image dataset used in this work consisted of 486,000 protein crystallography X-ray images.
Each image was manually annotated as either: 
(1) bad drop, 
(2) clear, 
(3) heavy precipitate, 
(4) large crystals, 
(5) light precipitate, 
(6) medium crystals, 
(7) micro crystals, 
(8) needles \& plates, 
(9) phase separation, 
and (10) small crystals.
This 10-way classification of images was adopted from the literature~\cite{sigdel2017feature,luft2011s}, and especially~\cite{luft2011s}, where it is concluded that a 10-way identification of the images is a better representation of the image content than a 2-way (crystal vs. non-crystal) one.

\begin{figure}[t]
\begin{center}
   \includegraphics[width=1\linewidth]{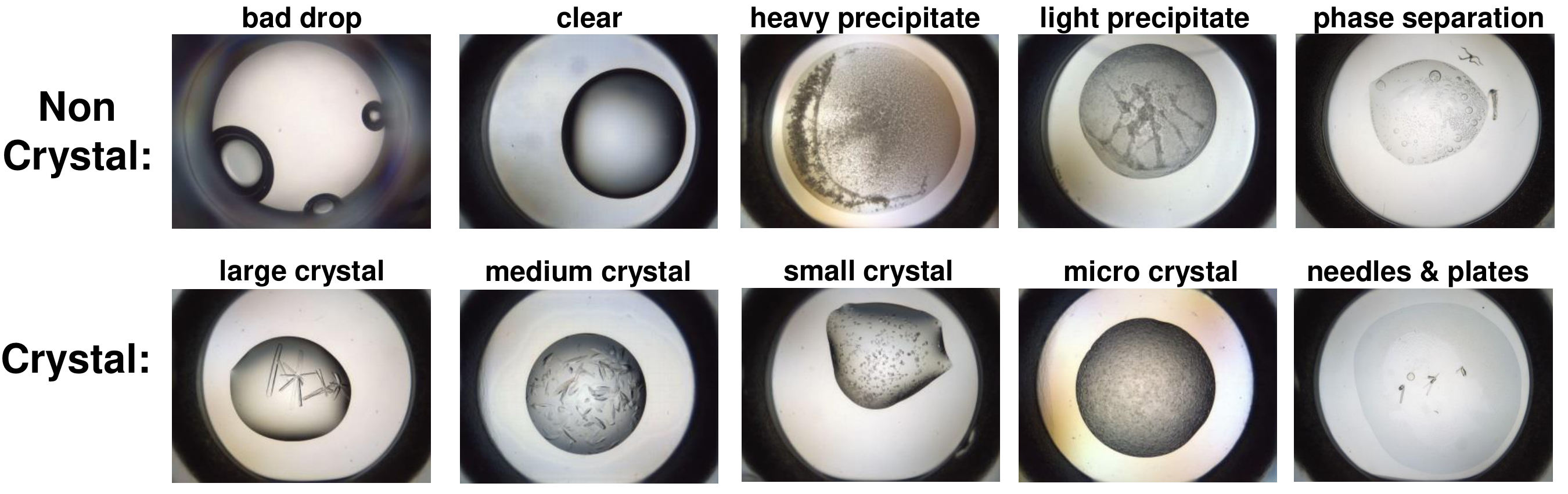}
\end{center}
\caption{
The images were assigned to ten labels based on their content.
Half of the labels denote the existence of crystals and the other half lack thereof.
The training and testing of the neural networks was based on the 10-way labeling of the data.
}
\label{fig:labels}
\end{figure}

As seen in Figure~\ref{fig:labels}, each of these categories implies the existence or lack thereof crystals.
The crystal categories denote the observation of crystal droplets formed in the crystallization drop. 
Large, medium, small, and micro crystals pertain to crystals of different sizes, while needles \& plates images contain protein crystals with a disc- or pin-like shape.
Each of the non-crystal categories reflect failed experiments, in which either the protein and the precipitant drop did not mix (bad drop), or the crystallization drop was missing altogether (clear).
In precipitate images, the protein forms grains rather than crystal and in the phase separation group liquid bubbles are observed in the crystallization drops.

\subsection{Data imbalance:}

\begin{figure}[t]
\begin{center}
   \includegraphics[width=0.53\linewidth]{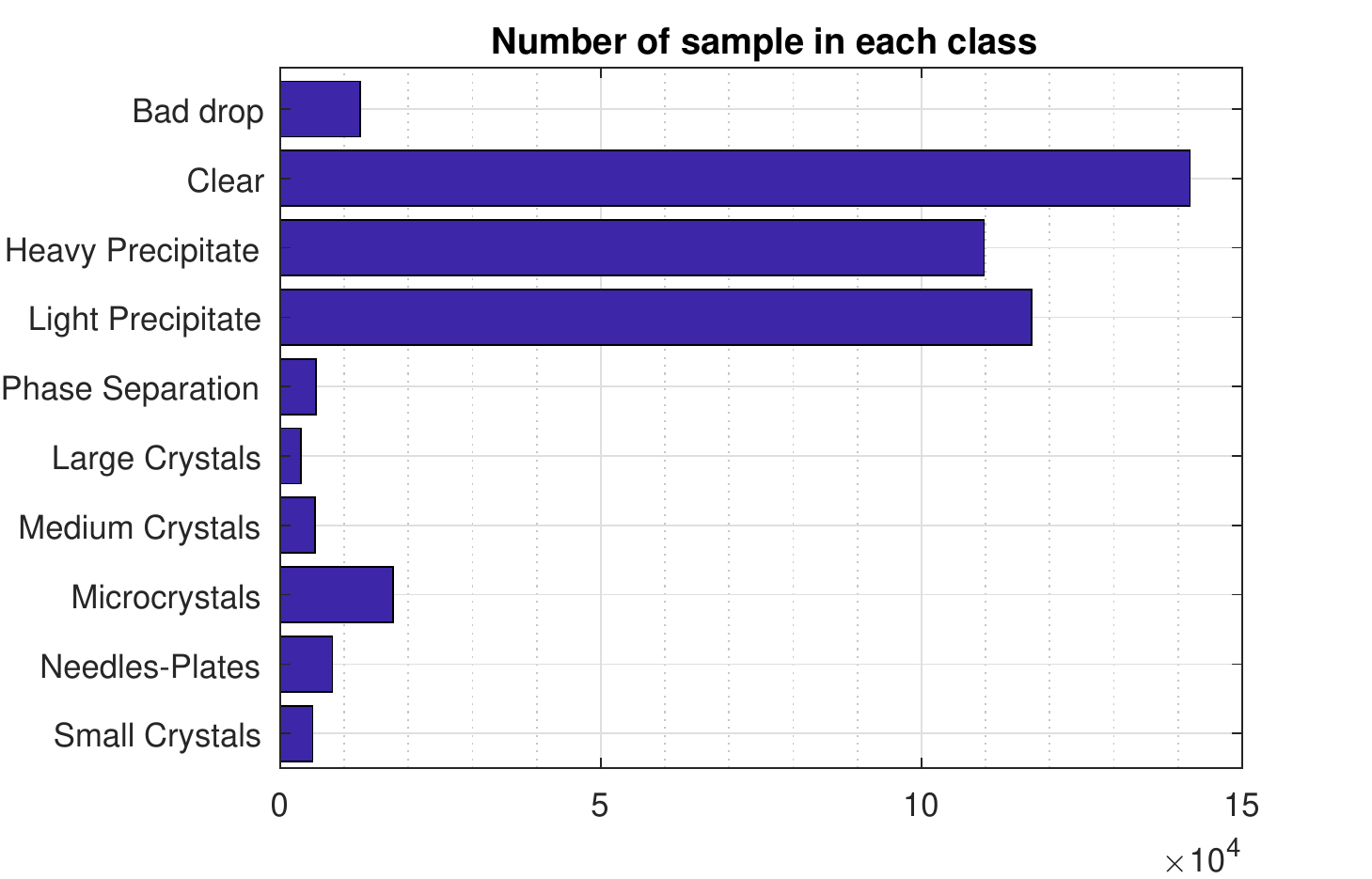}
\end{center}
\caption{
The low success rate of protein crystallization means that the resulting image datasets are significantly unbalanced towards non-crystal groups.
In our dataset, there were 9 times more non-crystal images than crystals.
A rebalancing strategy is therefore necessary to avoid the negative effects of data imbalance on classification results.
}
\label{fig:unbalanced}
\end{figure}

An intrinsic feature of protein crystallization images is data imbalance.
The low success rate of protein crystallization means that the crystal labels constitute only a small fraction of the dataset. 
In our case, only 9\% of the images in the total dataset contained crystals (Figure~\ref{fig:unbalanced}).
This imbalance is detrimental to the performance of the neural network.
During training, the network does not learn to recognize smaller classes due to its limited exposure to them.
This results in a large number of lost crystals, the detection of which is the ultimate goal of protein crystallization trials.
To make matters worse, this low performance, remains hidden during the tests, because the smaller number of the crystal groups stifles their effect on the testing accuracy.
Therefore, not addressing the imbalance issue in the dataset might result in models with seemingly small testing errors but low performance.

To remedy the adverse effect of data imbalance, we incorporated a data augmentation strategy in our image preprocessing pipeline.
Each original trial image was RGB with a size of 1280x960 pixels. 
Since colors did not show a visible indication of image content, all of the images were transformed to grayscale. 
The images were then divided into training, validation and testing datasets, where each of the groups made up  80\%, 5\%, and 15\% of the original dataset respectively.  
Thereafter, each image was randomly rotated, and then flipped horizontally and vertically, each with a probability of 0.5.
A 960x960-pixel window was then cropped from the center and then downsampled to 128x128 pixels to avoid lengthy training times.
Using this procedure, several images could be generated from each original image.
The number of generated images was inversely proportional to the size of the image group so that the resulting dataset was balanced.
The resulting dataset included 1,348,957 images with 1,076,552, 67,604, and 204,801 images for training, validation, and testing.
Although such synthetically augmented data is not as ideal as having a balanced dataset in the first place, it is preferable to discarding large portions of images from larger categories, which might result in a dataset too small for deep CNNs, which rely on big data for training.

\subsection{Neural Network Architectures}

We studied the performance of most successful, or relevant deep CNN architectures: CrystalNet~\cite{CrystalNet}, Lean CrystalNet\ifanonymous \hspace{.1cm} [anonymous] \else~\cite{GhafurianISBI}\fi, AlexNet~\cite{AlexNet}, VGG~\cite{VGG}, Inception version 3~\cite{Inception-V3,GoogLeNet}, and ResNet~\cite{ResNet}.
To the best of our information, CrystalNet was the only deep CNN architecture used for protein crystal detection.
It consists of four convolutional and three FC layers and was reported to achieve 90.8\% accuracy on the dataset introduced in~\cite{snell2008establishing1} and~\cite{snell2008establishing2}.
Looking for faster training times, we modified CrystalNet by adding one convolution, two pooling, and one FC layer, which resulted in a significant reduction in the number parameters while improving the classification accuracy.
We called this structure lean CrystalNet (LCN) to honor the original architecture and also the smaller computational cost.
A detailed description of LCN and its performance could be found in\ifanonymous \hspace{.1cm} [anonymous] \else~\cite{GhafurianISBI}\fi. 
The rest of the architectures used in this work are among the most successful network architectures in computer vision, which have not been used for the protein crystal detection problem.

For the optimization of the network parameters during training, we used mini-batch stochastic gradient with momentum~\cite{buduma2017fundamentals}.
The batch size was experimentally chosen as 64.
The training continued for 70 epochs and the value of the learning rate was divided by 10 every 20 epochs.
The validation error was measured after each epoch, and at the end of the training phase, the set of weight parameters which generated the best validation error during training was chosen as the training result.
The weights were initialized using the Guassian distribution function.
The implementation was done using TensorFlow~\cite{tensorflow} and TFLearn~\cite{tflearn2016}, and was run on an NVIDIA Quadro P6000 graphic processing unit.

\section{Results and Discussion}

The training and testing results can be seen in Table~\ref{tab:accuracies}.
The highest performing CNNs were ResNet-56, ResNet-32, and Inception-V3, with respective testing accuracy of 81.4\%, 80.57\%, and 79.40\%.
VGG came next with a testing accuracy of 79.39\%.
LCN showed a 4.48\% improvement compared to CrystalNet thanks to the added depth.
LCN also outperformed AlexNet and reached a top-1 accuracy 2.77\% better than AlexNet.
AlexNet and LCN have the same number of convolutional layers, but LCN has one additional FC layer, which must have contributed to better performance despite the larger number of features in AlexNet.

\begin{table}
\setlength{\tabcolsep}{12pt}
\centering
\begin{tabular}[1.0\textwidth]{|l|c|c|c|c|}
\hline
Network & Val. & \multicolumn{3}{c|}{Testing Accuracy } \\
\hhline{~~---}
Architecture & Acc. & top-1 & top-2 & top-3 \\
\hline\hline
CrystalNet  & 73.72\% & 74.16\% & 88.18\% & 93.35\% \\
LCN         & 77.41\% & 78.64\% & 92.24\% & 95.83\% \\
AlexNet     & 74.88\% & 75.87\% & 91.65\% & 96.18\% \\
VGG-16      & 78.64\% & 79.39\% & 92.49\% & 96.18\% \\
VGG-19      & 77.89\% & 78.70\% & 91.91\% & 95.76\% \\
Inception-V3& 79.40\% & 79.57\% & 94.13\% & 97.47\% \\
ResNet-32   & 80.57\% & 80.98\% & 95.47\% & 98.77\% \\
ResNet-56   & 81.43\% & 81.40\% & 95.94\% & 98.85\% \\
\hline
\end{tabular}
\caption{
Validation and testing results across the examined network architectures.
The validation column represents the best validation results during the training.
}
\label{tab:accuracies}
\end{table}

The better performance of more sophisticated CNNs supports our premise that protein crystal detection could benefit from higher level features and added depth.
The best performing architecture in our experiments, ResNet-56, was also the deepest one.
The 0.42\% improvement of ResNet by adding 24 layers to the original 32 ones suggests that even more accuracy could be gained by increasing depth.
However, unlike the ResNet, the addition of three more layers to VGG resulted in a 0.69\% drop in accuracy. 
Although the large number of parameters in VGG make it prone to overfitting, the lack of deterioration in validation accuracy during training suggests that this performance drop is the result of a degraded optimization due to the added complexity, rather than overfitting.
The authors of ResNet discuss this phenomenon, namely the degradation problem, in ~\cite{ResNet} and demonstrate that, due to the use of residual blocks, ResNet has a higher capacity to avoid this depth-related degradation.
Our experiments support their thesis by showing the improvement of results in ResNet vs. the deterioration in VGG when adding layers.

Using the top-n error is a common way of measuring classifier performance\cite{AlexNet,VGG,GoogLeNet,ResNet}.
This strategy is additionally useful in protein crystal detection for reducing the number of undetected crystals, in that an image is labeled as crystal if one of the labels with top-n output activation values is a crystal label.
The top-n testing accuracy for $n=\{1,2,3\}$ could be found in Table~\ref{tab:accuracies}.
To calculate the top-n accuracy of a model, the classification of an image is considered correct, if the true label of the image exists in the labels with $n$ maximum output layer activation.
The top-2 error was on average 14.2\% lower than top-1. 
And, the top-3 error was 3.8\% lower than top-2.
The best performing architectures using this measure were still ResNet and Inception-V3.
In ResNet-56 the true class of the image was among the top two in 95.94\% of the test cases.
Using a top-2 classification strategy decreases the rate of undetected crystals to 0.8\% from 6\% in top-1.
Using a top-3 strategy decreases the rate to less than 0.1\%.

\begin{figure}[t]
\begin{center}
   \includegraphics[width=0.5\linewidth]{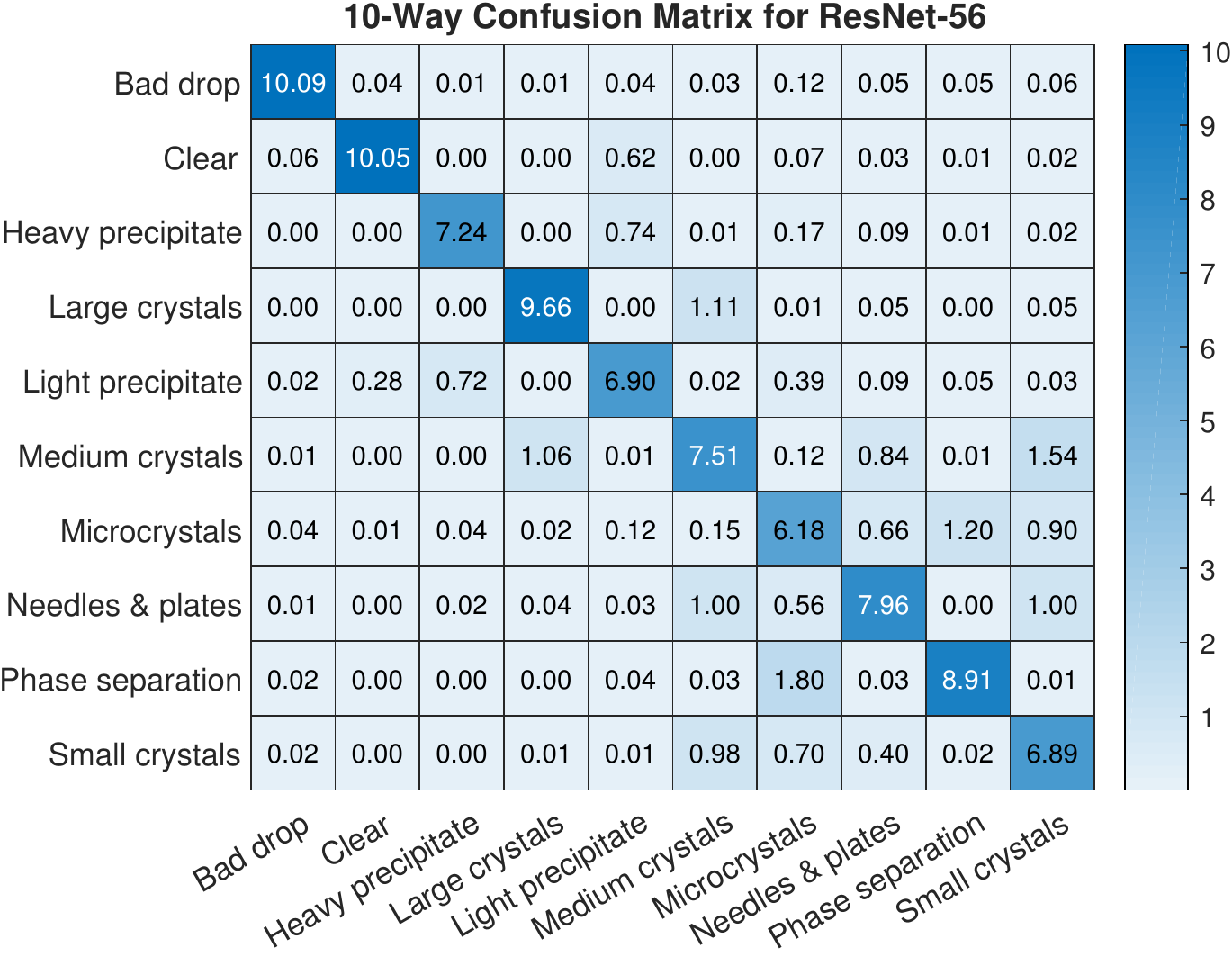}
\end{center}
\caption{
The confusion matrix for the 10-way classification of the test images using ResNet-56. 
Each number in the cells shows the percentage of the images, which belonged to the class associated with the column and were classified with the label associated with the row.
}
\label{fig:tenWay}
\end{figure}

The classification accuracy of the networks for each image label can be found in Table~\ref{tab:class-architecture}, along with the area under curve (AUC) computed from the receiver operating characteristic (ROC) plot (Table~\ref{tab:AUCs}).
Bad drop and clear images were the easiest to detect across different architectures with an average classification error of 3.07\% and 4.46\% respectively.
The label with consistently lowest accuracy among the architectures was the microcrystal.
On average, 42.65\% of the microcrystal images were misclassified.
A closer look at the confusion matrix of the results (Figure~\ref{fig:tenWay}) shows that the main source of this error is a confusion between microcrystal and phase separation images:
Approximately, half of microcrystal images were misclassified as phase separation.
Since phase separation images do not include crystals, this misclassification is unfavorable for the protein crystal detection application as it will lead to undetected crystals.
However, the cause of this confusion is understandable. 
Phase separation images include small liquid bubbles, which might resemble tiny crystals and fire similar neurons as microcrystals.

\begin{table*}[t]
\setlength{\tabcolsep}{3pt}
\centering
\begin{tabular}[1.0\textwidth]{|l|c|c|c|c|c|c|c|c|c|}
\hline
Architecture 
& \rotatebox[origin=c]{90}{ ResNet-56 }
& \rotatebox[origin=c]{90}{ ResNet-32 }
& \rotatebox[origin=c]{90}{ Inception-V3 }
& \rotatebox[origin=c]{90}{ VGG-19 }
& \rotatebox[origin=c]{90}{ VGG-16 }
& \rotatebox[origin=c]{90}{ AlexNet }
& \rotatebox[origin=c]{90}{ LCN }
& \rotatebox[origin=c]{90}{ CrystalNet }    
& \rotatebox[origin=c]{90}{ Average }  
\\
\hline\hline
Bad Drop            & 98.1\% & 97.6\% & 97.7\% &	97.2\%	& 97.2\% & 95.8\% & 96.1\% & 95.3\% & 96.9\% \\
Clear               & 96.6\% & 96.6\% & 95.7\% &	96.5\%	& 96.9\% & 92.6\% & 94.0\% & 95.1\% & 95.5\% \\
Heavy Precipit.     & 89.9\% & 90.7\% & 89.6\% &	91.8\%	& 91.9\% & 91.1\% & 88.8\% & 90.1\% & 90.5\% \\
Large Crystal       & 89.3\% & 88.7\% & 85.3\% &	82.3\%	& 83.6\% & 84.0\% & 84.3\% & 78.6\% & 84.5\% \\
Light Precipit.     & 81.0\% & 80.4\% & 79.2\% &	79.4\%	& 78.5\% & 73.4\% & 79.7\% & 75.5\% & 78.4\% \\
Medium Crystal      & 69.2\% & 69.8\% & 68.6\% &	68.8\%	& 67.9\% & 62.8\% & 68.8\% & 62.2\% & 67.3\% \\
Micro Crystal       & 61.1\% & 56.2\% & 63.0\% &    52.9\%	& 54.7\% & 54.1\% & 67.2\% & 49.3\% & 57.3\% \\
Needles \& Plates   & 78.0\% & 76.0\% & 78.4\% &	70.3\%	& 72.6\% & 71.9\% & 75.7\% & 66.3\% & 73.6\% \\
Phase Separat.      & 86.8\% & 88.1\% & 79.4\% &	85.3\%	& 87.0\% & 77.2\% & 67.9\% & 77.4\% & 81.1\% \\
Small Crystal       & 65.4\% & 67.1\% & 60.7\% &	64.9\%	& 65.9\% & 58.4\% & 64.9\% & 55.1\% & 62.8\% \\
\hline\hline
Class Average       & 81.5\% & 81.1\% & 79.8\% &	79.0\%  & 79.6\% & 76.1\% & 78.7\% & 74.5\% &        \\
\hline
\end{tabular}
\caption{
Classification accuracy for different classes across the studied architectures.
The microcrystal class had the lowest accuracy across all architectures.
}
\label{tab:class-architecture}
\end{table*}

\begin{table*}[t]
\setlength{\tabcolsep}{3pt}
\centering
\begin{tabular}[1.0\textwidth]{|l|c|c|c|c|c|c|c|c|c|c|}
\hline
& \rotatebox[origin=c]{90}{ Bad Drop }
& \rotatebox[origin=c]{90}{ Clear }
& \rotatebox[origin=c]{90}{ Heavy Pr. }
& \rotatebox[origin=c]{90}{ Large Cr. }
& \rotatebox[origin=c]{90}{ Light Pr. }
& \rotatebox[origin=c]{90}{ Med. Cr. }
& \rotatebox[origin=c]{90}{ Micro Cr. }
& \rotatebox[origin=c]{90}{ N. \& P. }    
& \rotatebox[origin=c]{90}{ Phase S. }
& \rotatebox[origin=c]{90}{ Small Cr. }  
\\
\hline\hline
ResNet-56       & 0.999 & 0.999 & 0.996 & 0.995 & 0.989 & 0.962 & 0.944 & 0.977 & 0.989 & 0.966 \\
ResNet-32       & 0.999 & 0.998 & 0.996 & 0.995 & 0.989 & 0.961 & 0.941 & 0.974 & 0.989 & 0.964 \\
Inception-V3    & 0.998 & 0.998 & 0.996 & 0.989 & 0.987 & 0.941 & 0.926 & 0.968 & 0.977 & 0.935 \\
\hline
\end{tabular}
\caption{
AUC values for the best performing architectures.
}
\label{tab:AUCs}
\end{table*}

The next most highly misclassified label was medium crystal with an average error of 67.31\%.
The misclassified medium crystal images were mostly labeled as needles \& plates, and large and small crystals.
Such misclassification is not detrimental for protein crystal detection purposes, because this confusion is between different crystal labels and therefore does not result in undetected crystals.
The cause of this error is that what separates medium crystals from small and large ones is sheer size.
The lack of a clear threshold makes it hard to tell these labels apart, even for a human annotator.
Also, needles \& plates are a special case of crystals with certain shapes, which makes them difficult to distinguish from other crystals at times.

\section{Conclusion}

Protein crystals are necessary for the study of protein molecule structures.
The empirical nature of this process necessitates the periodic observation of large numbers of trial images in search of forming crystals.
Although machine learning approaches have been used to solve this problem, the accuracy has not been high enough for them to replace manual inspection.
In this paper, we used six prominent deep CNN architectures for the classification of protein crystallization X-ray images and compared them to two simpler ones form the literature.
We used an internal dataset of 486,000 images and used a data augmentation step to tackle the data imbalance, increasing the dataset to more than a million images.
Our results showed that significant improvement was obtained by using more sophisticated CNNs compared to the existing ones.
The best performing CNN was ResNet with a 10-way accuracy of 81.43\%.
The rate of missed crystals for this model was 5.9\%, which could be reduced to 0.8\% and 0.0\% by using top-2 and top-3 detection strategies, with practically small number of false positives.
This level of precision makes it possible to automate crystal detection, freeing considerable time for crystallographers.

In our results, the most significant source of classification error was the mislabeling of microcrystal images as phase separation. 
This is due to the tiny liquid bubbles in phase separation images, which might trigger similar features to microcrystals.
Addressing this error in the future will improve protein crystal detection, because phase separation images do not hold crystals.
Therefore, eliminating this error will decrease the number of missed crystals as well.
Another source of error was the confusion between different crystal classes, which is expected as the boundaries are not explicit even for human annotators.
This error does not exacerbate detection results as it does not result in missed crystals.
Future work use the gained classification accuracy in this work for the design of a user interface to employ the trained models for compound discovery.

\bibliographystyle{splncs03}
\bibliography{bibliography}

\begin{thebibliography}{10}
\providecommand{\url}[1]{\texttt{#1}}
\providecommand{\urlprefix}{URL }

\bibitem{tensorflow}
Abadi, M., Agarwal, A., Barham, P., Brevdo, E., Chen, Z., Citro, C., Corrado,
  G.S., Davis, A., Dean, J., Devin, M., et~al.: Tensorflow: Large-scale machine
  learning on heterogeneous distributed systems. arXiv preprint
  arXiv:1603.04467  (2016)

\bibitem{bern2004automatic}
Bern, M., Goldberg, D., Stevens, R.C., Kuhn, P.: Automatic classification of
  protein crystallization images using a curve-tracking algorithm. Journal of
  applied crystallography  37(2),  279--287 (2004)

\bibitem{buduma2017fundamentals}
Buduma, N., Locascio, N.: Fundamentals of Deep Learning: Designing
  Next-generation Machine Intelligence Algorithms. " O'Reilly Media, Inc."
  (2017)

\bibitem{cumbaa2010protein}
Cumbaa, C.A., Jurisica, I.: Protein crystallization analysis on the world
  community grid. Journal of structural and functional genomics  11(1),  61--69
  (2010)

\bibitem{tflearn2016}
Damien, A., et~al.: Tflearn. \url{https://github.com/tflearn/tflearn} (2016)

\bibitem{GhafurianISBI}
Ghafurian, S., Orth, P., Strickland, C., Su, H., Patel, S., Soisson, S.,
  Dogdas, B.: Automated classification of protein crystallization x-ray images
  using deep convolutional neural networks. In: Biomedical Imaging (ISBI), 2018
  IEEE 15th International Symposium on. p. submitted. IEEE (2018)

\bibitem{ResNet}
He, K., Zhang, X., Ren, S., Sun, J.: Deep residual learning for image
  recognition. In: Proceedings of the IEEE conference on computer vision and
  pattern recognition. pp. 770--778 (2016)

\bibitem{hung2014protein}
Hung, J., Collins, J., Weldetsion, M., Newland, O., Chiang, E., Guerrero, S.,
  Okada, K.: Protein crystallization image classification with elastic net. In:
  Medical Imaging: Image Processing. p. 90341X (2014)

\bibitem{AlexNet}
Krizhevsky, A., Sutskever, I., Hinton, G.E.: Imagenet classification with deep
  convolutional neural networks. In: Advances in neural information processing
  systems. pp. 1097--1105 (2012)

\bibitem{liu2008image}
Liu, R., Freund, Y., Spraggon, G.: Image-based crystal detection: a
  machine-learning approach. Acta Crystallographica Section D: Biological
  Crystallography  64(12),  1187--1195 (2008)

\bibitem{luft2007efficient}
Luft, J.R., Wolfley, J.R., Said, M.I., Nagel, R.M., Lauricella, A.M., Smith,
  J.L., Thayer, M.H., Veatch, C.K., Snell, E.H., Malkowski, M.G., et~al.:
  Efficient optimization of crystallization conditions by manipulation of drop
  volume ratio and temperature. Protein science  16(4),  715--722 (2007)

\bibitem{luft2011s}
Luft, J.R., Wolfley, J.R., Snell, E.H.: What’s in a drop? correlating
  observations and outcomes to guide macromolecular crystallization
  experiments. Crystal growth \& design  11(3),  651--663 (2011)

\bibitem{rhodes2010crystallography}
Rhodes, G.: Crystallography made crystal clear: a guide for users of
  macromolecular models. Academic press (2010)

\bibitem{sigdel2017feature}
Sigdel, M., Dinc, I., Sigdel, M.S., Dinc, S., Pusey, M.L., Aygun, R.S.: Feature
  analysis for classification of trace fluorescent labeled protein
  crystallization images. BioData mining  10(1), ~14 (2017)

\bibitem{sigdel2013real}
Sigdel, M., Pusey, M.L., Aygun, R.S.: Real-time protein crystallization image
  acquisition and classification system. Crystal growth \& design  13(7),
  2728--2736 (2013)

\bibitem{VGG}
Simonyan, K., Zisserman, A.: Very deep convolutional networks for large-scale
  image recognition. arXiv preprint arXiv:1409.1556  (2014)

\bibitem{snell2008establishing2}
Snell, E.H., Lauricella, A.M., Potter, S.A., Luft, J.R., Gulde, S.M., Collins,
  R.J., Franks, G., Malkowski, M.G., Cumbaa, C., Jurisica, I., et~al.:
  Establishing a training set through the visual analysis of crystallization
  trials. part ii: crystal examples. Acta Crystallographica Section D:
  Biological Crystallography  64(11),  1131--1137 (2008)

\bibitem{snell2008establishing1}
Snell, E.H., Luft, J.R., Potter, S.A., Lauricella, A.M., Gulde, S.M.,
  Malkowski, M.G., Koszelak-Rosenblum, M., Said, M.I., Smith, J.L., Veatch,
  C.K., et~al.: Establishing a training set through the visual analysis of
  crystallization trials. part i:~ 150 000 images. Acta Crystallographica
  Section D: Biological Crystallography  64(11),  1123--1130 (2008)

\bibitem{GoogLeNet}
Szegedy, C., Liu, W., Jia, Y., Sermanet, P., Reed, S., Anguelov, D., Erhan, D.,
  Vanhoucke, V., Rabinovich, A.: Going deeper with convolutions. In:
  Proceedings of the IEEE conference on computer vision and pattern
  recognition. pp. 1--9 (2015)

\bibitem{Inception-V3}
Szegedy, C., Vanhoucke, V., Ioffe, S., Shlens, J., Wojna, Z.: Rethinking the
  inception architecture for computer vision. In: Proceedings of the IEEE
  Conference on Computer Vision and Pattern Recognition. pp. 2818--2826 (2016)

\bibitem{CrystalNet}
Yann, M.L.J., Tang, Y.: Learning deep convolutional neural networks for x-ray
  protein crystallization image analysis. In: AAAI. pp. 1373--1379 (2016)

\end{thebibliography}

\end{document}